\documentclass[11pt]{article}

\usepackage[preprint]{acl}

\usepackage{times}
\usepackage{latexsym}

\usepackage[T1]{fontenc}

\usepackage[utf8]{inputenc}
\usepackage{tcolorbox}
\usepackage{microtype}

\usepackage{inconsolata}

\usepackage{graphicx}
\usepackage{booktabs}
\usepackage{multirow}
\usepackage{graphicx}

\usepackage{amsmath}
\usepackage{amssymb}
\usepackage{mathtools}
\usepackage{amsthm}

\usepackage{multirow}
\usepackage[table,xcdraw]{xcolor}
\usepackage{tikz}
\usetikzlibrary{arrows.meta,calc,positioning,fit}
\usepackage[normalem]{ulem}
\useunder{\uline}{\ul}{}
\usepackage[capitalize,noabbrev]{cleveref}
\tcbuselibrary{breakable}
\theoremstyle{plain}

\theoremstyle{definition}

\theoremstyle{remark}

\usepackage{xurl}
\usepackage{hyperref}

\usepackage{algorithm}
\usepackage{algorithmic}
\usepackage{enumitem}

\title{Adaptive Multi-Resolution Procedural Knowledge Compression for Large Language Models}

\usepackage{authblk}

\author[1]{\textbf{Changyue Wang}\thanks{cy-wang24@mails.tsinghua.edu.cn}}
\author[1]{\textbf{Weihang Su}}
\author[1]{\textbf{Qingyao Ai}\thanks{Corresponding Author: aiqy@tsinghua.edu.cn}}
\author[1]{\textbf{Yichen Tang}}
\author[1]{\protect\\\textbf{Runzhong Qiao}}
\author[1]{\textbf{Xuancheng Li}}
\author[1]{\textbf{Min Zhang}}
\author[1]{\textbf{Yiqun Liu}}

\affil[1]{Department of Computer Science and Technology, Tsinghua University}

\begin{document}
	\maketitle
	\begin{abstract}
		Large language models (LLMs) are widely used to tackle complex tasks with autonomous workflows.
		Recently, reusable natural language skills have emerged as a popular paradigm to inject procedural knowledge into LLM applications.
		Since popular skills are often invoked repeatedly, placing their full text in every context significantly increases prefill cost and latency.
		While text compression techniques have the potential to solve this problem, most existing methods are designed to compress factual knowledge in documents instead of procedural knowledge, making them insufficient for skill compression.
		In this paper, we argue that an effective skill compression method should: 1) preserve logical dependencies among workflows and tool protocols,
		2) enable lightweight, offline compression for frequently updated community skills, and 3) be adaptable to varying complexities across skills.
		To address this, we present \textbf{\textsc{SKIM}} (\underline{SKI}ll co\underline{M}pression), an adaptive multi-resolution soft token compression framework for procedural skills.
		Depending on the complexity of each skill, \textsc{SKIM} creates different numbers of soft tokens that not only improve the efficiency of LLM inference, but also preserve the effectiveness of skill usage.
		Experiments indicate that \textsc{SKIM} compresses skills to 30 to 60 percent of their original token length while preserving task performance better than existing compression methods.\footnote{We have released our code at \url{https://github.com/bebr2/SKIM}}
	\end{abstract}
	
	\section{Introduction}

	Skill, a type of data structure that stores reusable procedural knowledge specifying when and how a large language model (LLM) should invoke a capability, follow a workflow, or interact with tools, has been shown to be effective in improving the performance of LLM agents in downstream tasks.
	Such skills are often written as natural-language files such as \texttt{SKILL.md}, making them easy for users to create, edit, and share~\citep{zhou2026comprehensivesurveyagentskills, su2026skillretrievalaugmentationagentic}.
	While this natural-language format has made skills an attractive abstraction for building, distributing, and reusing agent capabilities, deploying them at scale introduces substantial costs in terms of token consumption.
	Typically, every invocation requires placing the full skill text into the model's context window. As agent skill ecosystems like OpenClaw \cite{openclawOpenClawPersonal} increasingly power daily workflows for many developers, popular skills are downloaded and then invoked across diverse systems \citep{jiang2026harmfulskillbenchharmfulskillsweaponize}, causing redundant token overhead to accumulate.
	Therefore, even reducing a small fraction of prompt tokens for these skills can translate into aggregate computational and latency savings.
	Furthermore, since users can freely modify these natural-language skills and deploy them across various inference backends, the direct application of traditional key-value (KV) cache reuse \citep{vllm} becomes unfeasible.

	A natural way to reduce this overhead is to compress skill content before deployment.
	Existing text compression methods can be broadly divided into hard compression, which selects discrete text tokens~\citep{pan-etal-2024-llmlingua,jiang-etal-2024-longllmlingua}, and soft compression, which maps contexts into continuous representations~\citep{icae,li-etal-2025-500xcompressor,atacompressor}.
	These methods have been primarily developed and evaluated for generic prompts, long-context understanding, or document-centric tasks, where preserving salient semantic content or query-relevant evidence is often sufficient.
	In such settings, a compressed context can remain useful as long as the critical facts or evidence are still recoverable.
	
\begin{figure}[!t]
		\centering
		\includegraphics[width=\columnwidth]{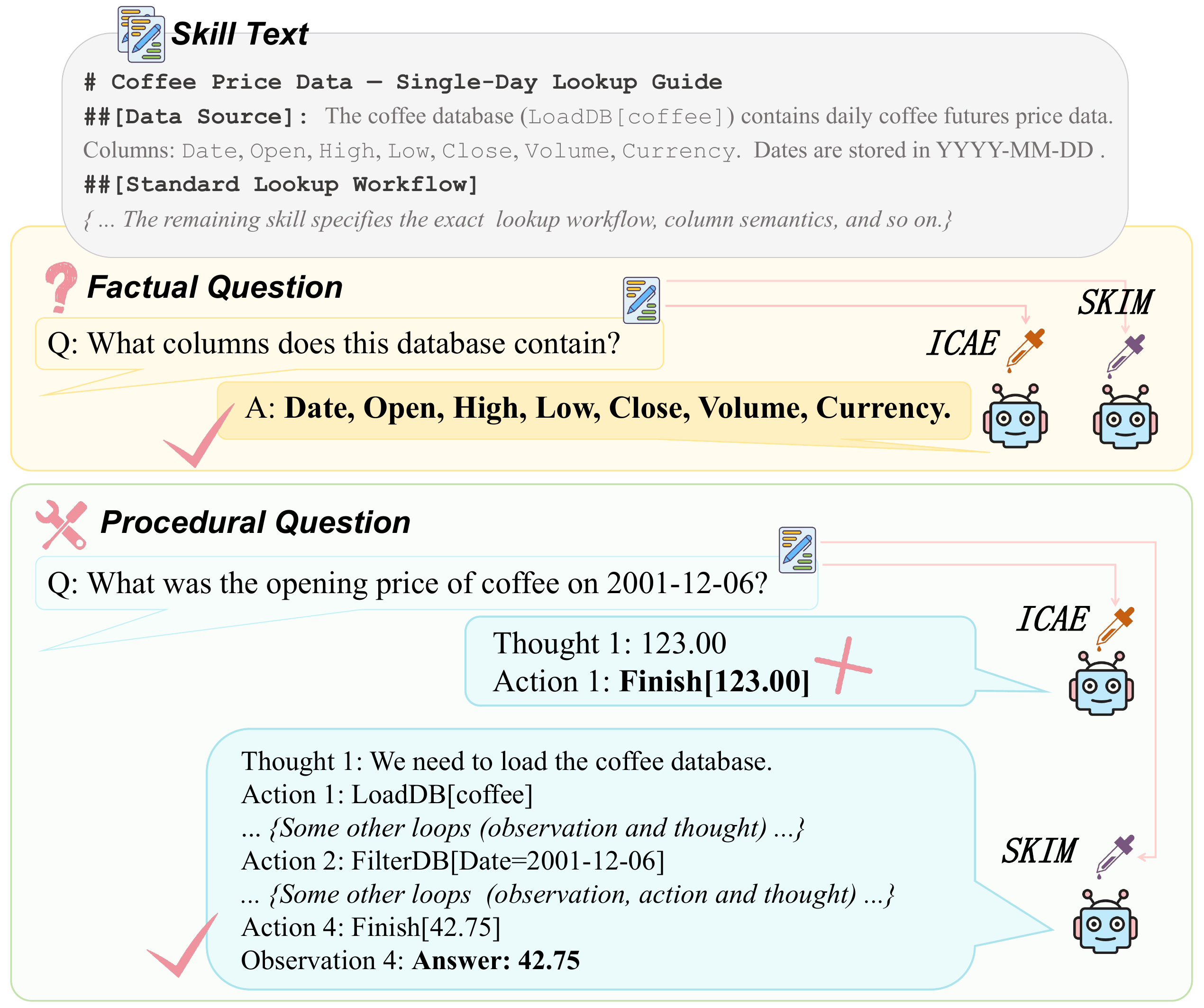}
		\caption{A ToolQA example. Factual information can be recovered from compressed skills, but the procedural question requires preserving workflow specified by skill.}
		\label{fig:case}
	\end{figure}

	Unfortunately, unlike descriptive documents, skills encapsulate procedural knowledge \citep{chen2026skvmrevisitinglanguagevm}, presenting three unique compression challenges for existing text compression methods.
	First, skill compression must preserve procedural dependencies.
	Such information is not stored as simply as some factual keywords.
	Therefore, rather than merely retaining isolated facts, compression must strictly preserve the dependencies between conditions, tool arguments, and other information in the skill, as severing even one logical link can lead to execution failure.
	For example, Figure \ref{fig:case} shows a case where we use ICAE \citep{icae}, a representative context compression method for LLMs, to compress a skill.
	When the compressed skill is used for a factual question, the performance of the system is fine.
	However, when it is used on a task that actually requires the procedural knowledge in the skill, ICAE breaks the system and produces wrong results.

	Second, the sharing and deployment paradigm of skills differs fundamentally from static texts \citep{hu2026redskillsblueskills}.
	Since users actively share, modify, and distribute skills, their compressed representations must be rapidly regenerable and easily transmittable.
	Consequently, the resulting artifacts must be storage efficient.
	Furthermore, the process of encoding a new or updated skill must be computationally lightweight, relying only on a simple forward pass rather than expensive online gradient backpropagation required by temporary training methods like TokMem \citep{wu2026tokmemonetokenproceduralmemory}.
	
	Third, skills differ significantly in complexity and information density.
	A fixed compression rate cannot simultaneously adapt to all skills, and the optimal rate for one skill might even vary across models.
	Therefore, an ideal framework should represent a skill under multiple token budgets and automatically select the appropriate compression resolution for specific skill and model pairs.

	To address these challenges, we propose \textbf{\textsc{SKIM}} (\underline{SKI}ll co\underline{M}pression), an adaptive multi-resolution soft token compression framework.
	Following dual model compression architectures such as DRIFT \citep{xie2026decoupledreasoningimplicitfact}, \textsc{SKIM} uses a compressor to encode the skill and a projector to map the resulting soft tokens into the target LLM space.
	\textsc{SKIM} is trained via a progressive three-stage paradigm, including skill reconstruction, procedural warm-up, and skill task alignment, to preserve executable dependencies.
	During skill task alignment, the target LLM is adapted via a single Low-Rank Adaptation (LoRA) \citep{lora} module, ensuring parameter efficiency.
	To accommodate varying skill complexities, \textsc{SKIM} supports representing a skill under multiple token budgets (resolutions) simultaneously.
	Before deployment, an offline self-judgment mechanism automatically selects the lowest-budget resolution that maintains execution accuracy.

	\begin{table*}[!t]
\centering
\resizebox{0.9\textwidth}{!}{%
\begin{tabular}{@{}llccc@{}}
\toprule
\textbf{Work}                                                        & \textbf{Form} & \textbf{Offline Compression} & \textbf{Lightweight} & \textbf{Knowledge Type} \\ \midrule
LLMLingua-2 \citep{pan-etal-2024-llmlingua}         & hard tokens   & Yes              & Yes                  & General       \\
LongLLMLingua \citep{jiang-etal-2024-longllmlingua} & hard tokens   & No               & Yes                  & Factual       \\
ICAE \citep{icae}                                   & soft tokens   & Yes              & Yes                  & Factual       \\
500xCompressor \citep{li-etal-2025-500xcompressor}  & soft tokens \& KV Cache & Yes & Yes & Factual    \\
DRIFT \citep{xie2026decoupledreasoningimplicitfact} & soft tokens   & No               & Yes                  & Factual       \\
TokMem \citep{wu2026tokmemonetokenproceduralmemory} & soft tokens       & Yes & No  & Procedural \\
\textsc{SKIM}                                      & soft tokens   & Yes              & Yes                  & Procedural    \\ \bottomrule
\end{tabular}%
}
\caption{Comparison of prompt compression methods. Lightweight denotes forward compression without gradient optimization. Knowledge Type specifies the form of knowledge targeted by each method.
}
\label{tab:related_work_comparison}
\end{table*}

	This design perfectly aligns with the nature of skill deployment paradigm.
	Developers can distribute skills as lightweight, pre-computed soft token bundles for different mainstream target models, functioning similarly to a prebuilt artifact in software distribution.
	When a skill is updated, its compressed representation is instantly regenerated via a single forward pass.
	Furthermore, the multi-resolution design seamlessly adapts to the varying complexities of different skills.
	During online inference, the serving infrastructure simply loads the pre-selected soft tokens and switches to the shared LoRA adapter, without additional latency.

	Our contributions are summarized as follows:
	\begin{itemize}[leftmargin=*]
		\item We formulate procedural skill compression as distinct from factual document compression, where preserving executable logic is central.
		\item We propose \textsc{SKIM}, a skill compression framework that combines progressive training with offline self-judgment for resolution selection.
		\item We evaluate \textsc{SKIM} on several skill-based datasets, showing that it outperforms existing methods in token reduction while maintaining high skill task accuracy.
	\end{itemize}

	\section{Related Work}

	\subsection{Skills for Large Language Models}

	Large Language Models (LLMs) have evolved into agentic problem solvers for complex tasks.
	Recent work has extended retrieval augmented generation from declarative knowledge to procedural capabilities, where reusable external resources are selected and applied at inference time \citep{su2024dragin,su2024mitigating,su2025parametric,su2025dynamic}.
	A growing paradigm is therefore to package procedural knowledge as \textit{skills} \citep{chen2026skvmrevisitinglanguagevm}, which can augment agent frameworks such as OpenClaw \citep{openclawOpenClawPersonal}.
	A skill is usually written as a \texttt{SKILL.md} file.
	During inference, after one or more skills are selected, their main body fields are loaded into the LLM context.
	A recent large scale study reports that online skills average more than 2,000 tokens and that some exceed 10,000 tokens \citep{cho2026skillretlargescalebenchmarkskill}.
	Moreover, platforms like ClawHub \citep{clawhubClawHub} provide centralized repositories for skill sharing, accelerating their adoption, which significantly increases prompt token consumption.
	
	\subsection{Prompt Compression} 
	
	Prompt compression generally reduces computation cost by distilling redundant context, falling into two categories.
	Hard compression explicitly prunes discrete tokens.
	For instance, LLMLingua-2 \citep{pan-etal-2024-llmlingua} provides task-agnostic compression, whereas LongLLMLingua \citep{jiang-etal-2024-longllmlingua} utilizes a query-aware mechanism for retrieval scenarios but introduces additional online latency.
	Furthermore, these hard techniques often discard structural elements necessary for executable logic.
	Alternatively, soft compression projects text into continuous representations.
	ICAE \citep{icae} maps text into continuous embeddings, and 500xCompressor \citep{li-etal-2025-500xcompressor} retains KV cache tensors.
	For this reason, 500xCompressor introduces massive storage pressure, rendering it unsuitable for skill distribution scenarios.
	To address the lack of procedural compression, TokMem \citep{wu2026tokmemonetokenproceduralmemory} distills an instruction sequence into a single token.
	However, it demands costly gradient-based optimization for every new procedure, making it impractical for skill ecosystems where community iterations are rapid.
	Recently, DRIFT \citep{xie2026decoupledreasoningimplicitfact} leverages a dual model soft token architecture to extract factual information.
	Yet, it requires online compression to process retrieved long texts and the user query, thereby introducing runtime overhead.
	Building upon such architecture, \textsc{SKIM} optimizes a compressor and a target LLM specifically for procedural skills, while ensuring offline compression and lightweight forward pass.
	We summarize these differences in Table \ref{tab:related_work_comparison}.
	
	\section{Methodology}
	
	In this section, we describe \textsc{SKIM} as a framework for replacing selected skill content with continuous tokens.
	We define the compression task, then present the architecture, the progressive training, offline resolution selection, and inference.
	
	\subsection{Overview}
	
	A skill often contains name, description, content, and other metadata fields.
	The content field is the main body and commonly specifies procedural guidance, tool descriptions, and usage constraints.
	We focus on the content field of a skill, since this field is the skill body loaded into the LLM context after selection.
	Given skill content $s$ and a user query $u$, the goal is to replace the full skill content text in the input prompt with a compact continuous representation while preserving the procedural behavior specified by $s$.
	For a target LLM $M$ with embedding dimension $d_M$, \textsc{SKIM} produces a matrix $E_K(s) \in \mathbb{R}^{K \times d_M}$ under a token budget $K$.
	The budget $K$ is selected from an ordered set $\mathcal{K}$, where smaller values correspond to more aggressive compression.
	\textsc{SKIM} contains a compressor, learnable slot tokens, a projector, and a target LLM adapted with LoRA.
	The compressor reads the skill  and slot tokens, the projector maps slot position hidden states into the target LLM embedding space, and the resulting soft tokens are inserted into the target context together with the user query.
	
\begin{figure*}[!t]
		\centering
		\includegraphics[width=\textwidth]{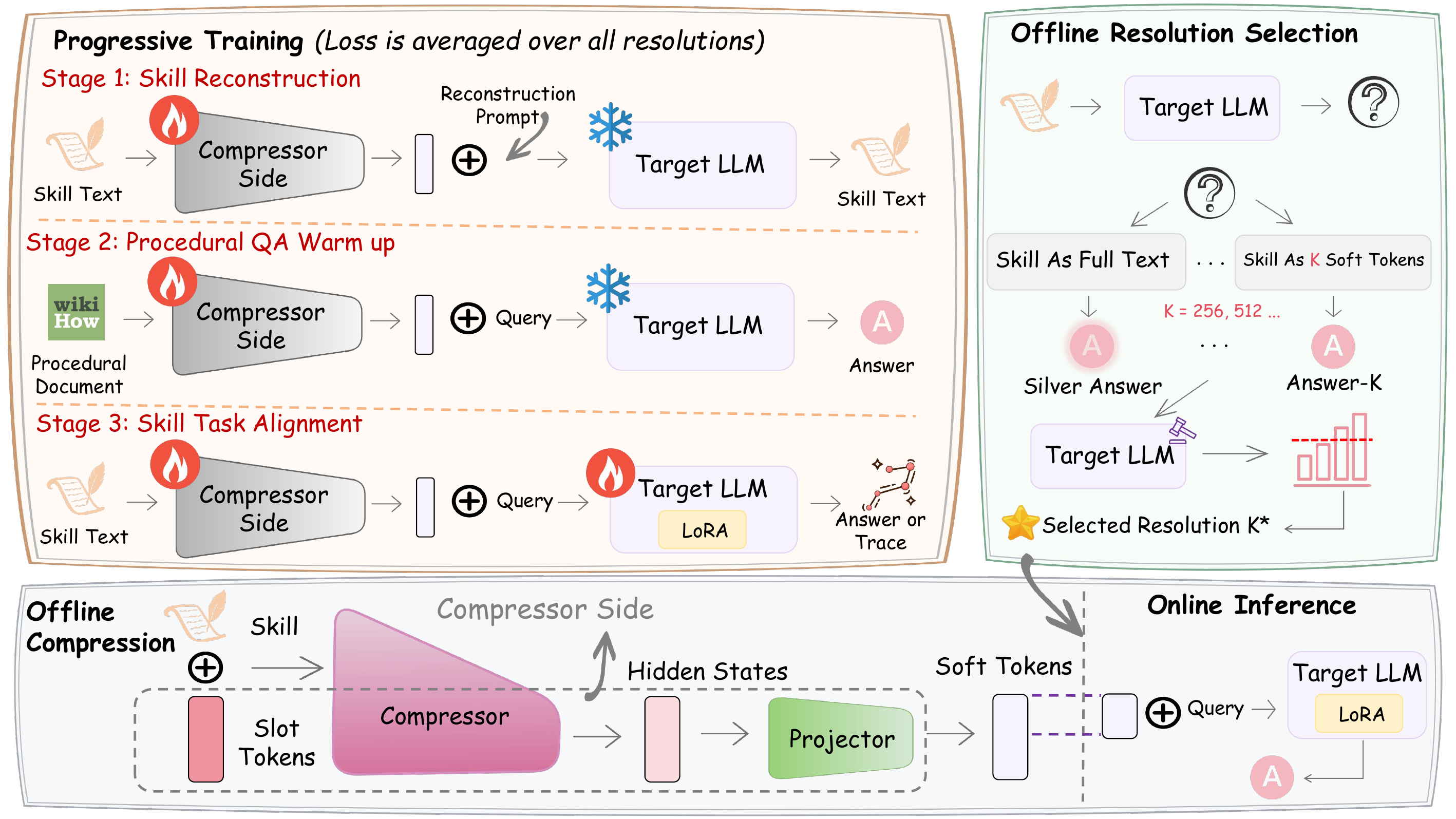}
		\caption{Overview of \textsc{SKIM}. The upper panel illustrates the training process and offline resolution selection, where the compressor (consisting of slot tokens, an LLM-based compressor, and an MLP projector) learns to convert skills into soft-token prefixes aligned with the target LLM's embedding space. The lower panel shows the inference architecture, where \textsc{SKIM} selects the soft-token prefix corresponding to the offline-determined optimal resolution and prepends it to the target LLM.}
		\label{fig:methodology_overview}
	\end{figure*}

	Figure \ref{fig:methodology_overview} summarizes the pipeline.
	During training, the same skill is optimized under multiple resolutions so that the compressor learns representations that remain useful when truncated.
	Before deployment, \textsc{SKIM} compresses each skill for each target model using the corresponding compressor, projects the representation into the target model space, and stores the full soft token artifact.
	It then runs an offline diagnostic evaluation to record the resolution selected for that skill and target model.
	At inference, the target LLM receives text tokens for the query and soft tokens for each skill, and switches to the LoRA adapter to generate response.
	
	\subsection{Multi-Resolution Compressor}
	
	\textsc{SKIM} follows the dual model soft token design used in recent context compression work \citep{liu2025contextcascadecompressionexploring, xie2026decoupledreasoningimplicitfact}. 
	The compressor $C_{\theta}$ is an autoregressive backbone LLM that receives tokenized skill content followed by $K_{\max}$ learnable slot tokens.
	These slot tokens are trained parameters and remain fixed after training.
	After the compressor forward pass, the hidden states at the slot token positions form a latent representation $Z(s) \in \mathbb{R}^{K_{\max} \times d_C}$, where $d_C$ is the compressor hidden dimension.
	A multilayer perceptron (MLP) projector $P_{\phi}$ maps this latent representation into the target model space:
	\begin{equation}
		E_{K_{\max}}(s) = P_{\phi}(Z(s)) \in \mathbb{R}^{K_{\max} \times d_M}.
	\end{equation}
	
	For a lower resolution $K < K_{\max}$, \textsc{SKIM} uses the prefix $E_K(s)=E_{K_{\max}}(s)_{1:K}$.
	Since one high resolution representation can be truncated to obtain lower resolutions, a platform can store a single artifact per skill and target model, then choose a smaller budget without recomputing the representation.
	When a prompt contains multiple skills, \textsc{SKIM} compresses each independently.
	The projected soft token sequences are then concatenated in the same order as the selected skills.
	Then, the target LLM with LoRA adapter is called with the soft token sequence and the user query.
	
	\subsection{Progressive Training}
	
	\textsc{SKIM} uses three training stages.
	Each stage optimizes the compressor, learnable slot tokens, and projector with full parameter updates, but differs in supervision and whether the target LLM is adapted.
	For each training example, the loss is averaged over resolutions returned by $\mathcal{K}$.
	This objective teaches the model to support multiple budgets.
	
	\paragraph{Stage 1: Skill Reconstruction.}
	The first stage learns a general skill representation from a large corpus of skills collected from the web.
	The target LLM is frozen.
	Given a skill $s$, \textsc{SKIM} inserts $E_K(s)$ into a reconstruction prompt and optimizes the negative log likelihood of the original text:
	\begin{equation}
		\mathcal{L}_{\mathrm{rec}} =
		- \frac{1}{|\mathcal{K}|}
		\sum_{K \in \mathcal{K}}
		\log p_M(s \mid E_K(s), r_{\mathrm{rec}}),
	\end{equation}
	where $r_{\mathrm{rec}}$ is a short reconstruction instruction.
	This stage encourages the slot states to retain the core information in the skill content before the model is exposed to downstream tasks.
	
	\paragraph{Stage 2: Procedural Question Answering Warm Up.}
	The second stage aligns the compressed representation with procedural question answering.
	Unlike factual knowledge Question Answering (QA), where labeled question answer pairs are relatively abundant, labeled QA data for executable skills are much more limited.
	To address this data scarcity, we introduce this procedural warm-up stage that repurposes WikiHow \citep{koupaee2018wikihowlargescaletext} as weakly supervised question answering data.
	WikiHow is a large scale summarization dataset crawled from ``how-to articles'' on the \textit{WikiHow} website.
	For our purpose, the article title serves as the question, the article body serves as procedural text, and the summary serves as a concise answer.
	Each example therefore contains a procedural document $d$, a query $q$, and an answer $a$.
	The document is compressed while the query remains plain text:
	\begin{equation}
		\mathcal{L}_{\mathrm{qa}} =
		- \frac{1}{|\mathcal{K}|}
		\sum_{K \in \mathcal{K}}
		\log p_M(a \mid E_K(d), q).
	\end{equation}
	Since the target LLM is frozen, this stage mainly trains the compressor side to expose information in a form that the target model can consume.
	The warm-up signal moves the compressor beyond reconstruction by aligning procedural text with answer generation before skill tasks are introduced.
	
	\paragraph{Stage 3: Skill Task Alignment.}

	The third stage uses the same task format and loss as the warm-up stage but replaces WikiHow documents with real skills from the first stage.
	Our goal here is to align the compressed continuous tokens and the target model's behavior with actual skill-conditioned tasks, which often require complex decision making, tool calls, or the joint use of multiple skills.
	To build this training data, we employ a high-capacity LLM (GPT-5.2 in our implementation \citep{openai2025introducinggpt52}) as an evaluator.
	This LLM first evaluates the collected skills, filtering out documents with weak procedural content.
	For the remaining skills, it generates skill-dependent user questions and extracts relevant metadata (e.g., tool requirements or splittability).
	Then, supervised answers are generated by the target model.
	We prompt it to follow skill guidance to prevent it from bypassing the skill text and relying solely on its parametric memory.
	Furthermore, to effectively train the model on complex real-world behaviors, our data construction specifically addresses two critical properties of skills:
	
	\begin{itemize}[leftmargin=*]
		\item \textbf{Tool Usage Simulation:} Many skills specify external tool usage. To ensure the framework can trigger correct tool workflows, we synthesize ReAct-style traces \citep{react} with interleaved Thought, Action, and Observation steps.
		Since accessing APIs for collected diverse web skills is impractical, when generating answers, we use the target LLM itself as a simulator to generate simulated tool observations conditioned on the tool schema, query, and partial trajectory.
		
		\item \textbf{Multi-Skill Composition:} Some queries require integrating multiple skills.
		Since evaluating every pairwise combination of skills is computationally prohibitive, we use a top-down ``skill split'' strategy.
		The high-capacity LLM rewrites a complex skill into self-contained subskills while using the original skill text to generate the corresponding question.
		During training, \textsc{SKIM} compresses these subskills independently and concatenates their soft tokens to model multi-skill scenarios.
	\end{itemize}
	
	Algorithm \ref{alg:stage3_data_construction} in Appendix \ref{app:stage3_data_construction} summarizes this data construction procedure.
	In this training stage, the target LLM is adapted through a LoRA module, while the compressor, slot tokens, and projector remain trainable.
	This single, unified LoRA adapter is shared across all skills, aligning the target model's behavior with the continuous skill tokens, without modifying its full parameters.

	\subsection{Offline Resolution Selection}
	
	Since skills differ in complexity, different skills require different compression budgets.
	A simple skill may be represented with only a few soft tokens, whereas a multi-step procedural skill may require a higher resolution.
	We therefore select a skill-model specific resolution before deployment.
	Algorithm \ref{alg:resolution_selection} shows the related pseudocode.
	
	For each skill $s$, \textsc{SKIM} selects its compression resolution through an offline calibration step.
	The target LLM first generates $N$ diagnostic questions from the original skill text.
	For each question, the target LLM is prompted with the full skill text to produce a reference answer, which serves as a silver standard because no human label is available.
	We then replace the full skill text with compressed soft tokens at each resolution $K \in \mathcal{K}$, generating one candidate answer per resolution.
	The model itself compares each candidate answer with the corresponding full-text reference answer and computes a fidelity score $\alpha_K(s)$ for each resolution.
	Given a threshold $\tau$, \textsc{SKIM} chooses the smallest resolution whose fidelity reaches the threshold:
	\begin{equation}
		K^{\star}(s) =
		\min \{K \in \mathcal{K}: \alpha_K(s) \geq \tau \}.
	\end{equation}
	If no compressed resolution reaches the threshold, the system falls back to the original skill text.
	This offline calibration adds no latency to user requests.
	
	\subsection{Inference}
	
	At inference time, a single skill uses its selected resolution $K^{\star}(s)$.
	For a prompt with multiple skills, \textsc{SKIM} compresses each skill and concatenates their soft token sequences as in the third training stage.
	The system fetches the corresponding soft token prefixes, activates the LoRA adapter, concatenates the soft tokens with the user query, and runs the target LLM forward pass to decode the final response.
	
	Importantly, the LoRA adapter trained in Stage 3 is a single, unified module shared across \textit{all} skills, including unseen ones.
	The online path merely involves skill lookup (which depends on the skill retrieval system), soft token loading, and standard LLM decoding with the LoRA activated.
	Modern inference engines like vLLM \citep{vllm} support pre-loading and switching LoRA adapters with minimal overhead.
	For queries not requiring skills, the system seamlessly deactivates the adapter to preserve the base model's original capabilities.

	When a skill is updated, the shared LoRA or the compressor requires no retraining.
	The publisher can regenerate the corresponding soft token artifact through a compressor forward pass,
	and may rerun the offline resolution selection step only when the edit is substantial.
	This preserves the convenient distribution property of natural language skills.
	
	\section{Experimental Setup}
	\label{sec:experimental_setup}

	\subsection{Tasks and Data}
	
	\begin{table*}[!t]
\centering
\scriptsize
\caption{Main results on the five datasets with golden skills (accuracy (\%) / average added skill tokens).
500xCom. denotes 500xCompressor.
\textsc{SKIM} Fix-256 and Fix-512 use fixed soft token budgets for every skill, while Adaptive selects between 256 tokens, 512 tokens, and the full text fallback with the offline exam.
LLMLingua-2 (S/M/L) use compression ratios chosen to roughly match the three \textsc{SKIM} token settings.
Bold \textsc{SKIM} entries mark cases where the method has higher accuracy and fewer tokens than at least one LLMLingua-2 baseline.}
\label{tab:main_results}
\begingroup
\renewcommand{\arraystretch}{1.0}
\setlength{\tabcolsep}{3.0pt}
\resizebox{0.95\textwidth}{!}{%
\begin{tabular}{lllcccccccccc}
\toprule
\multirow{2}{*}{\textbf{Model}} & \multirow{2}{*}{\textbf{Dataset}} & \multirow{2}{*}{\textbf{Metric}} & \multirow{2}{*}{\textbf{Naive}} & \multirow{2}{*}{\textbf{Full Text}} & \multicolumn{2}{c}{\textbf{Soft Baselines}} & \multicolumn{3}{c}{\textbf{LLMLingua-2}} & \multicolumn{3}{c}{\textbf{\textsc{SKIM}}} \\
\cmidrule(lr){6-7} \cmidrule(lr){8-10} \cmidrule(lr){11-13}
 & & & & & \textbf{500xCom.} & \textbf{ICAE} & \textbf{S} & \textbf{M} & \textbf{L} & \textbf{Fix-256} & \textbf{Fix-512} & \textbf{Adaptive} \\
\midrule
\multirow{10}{*}{\rotatebox[origin=c]{90}{Qwen3-8B}} & \multirow{2}{*}{BigCodeBench} & \textbf{Acc. (\%)} & 40.96 & 49.30 & 0.18 & 29.47 & 40.70 & 43.60 & 46.05 & \textbf{44.21} & \textbf{45.53} & \textbf{46.93} \\
 &  & \cellcolor[HTML]{F8F5FF}\textbf{\# Token} & \cellcolor[HTML]{F8F5FF}0 & \cellcolor[HTML]{F8F5FF}4676 & \cellcolor[HTML]{F8F5FF}256 & \cellcolor[HTML]{F8F5FF}256 & \cellcolor[HTML]{F8F5FF}1300 & \cellcolor[HTML]{F8F5FF}2655 & \cellcolor[HTML]{F8F5FF}3362 & \cellcolor[HTML]{F8F5FF}\textbf{708} & \cellcolor[HTML]{F8F5FF}\textbf{1414} & \cellcolor[HTML]{F8F5FF}\textbf{3172} \\
\cmidrule(lr){2-13}
 & \multirow{2}{*}{CHAMP} & \textbf{Acc. (\%)} & 48.88 & 68.16 & 3.14 & 27.80 & 52.47 & 55.61 & 57.40 & \textbf{54.26} & \textbf{61.43} & \textbf{65.92} \\
 &  & \cellcolor[HTML]{F8F5FF}\textbf{\# Token} & \cellcolor[HTML]{F8F5FF}0 & \cellcolor[HTML]{F8F5FF}1941 & \cellcolor[HTML]{F8F5FF}256 & \cellcolor[HTML]{F8F5FF}256 & \cellcolor[HTML]{F8F5FF}594 & \cellcolor[HTML]{F8F5FF}1122 & \cellcolor[HTML]{F8F5FF}1477 & \cellcolor[HTML]{F8F5FF}\textbf{438} & \cellcolor[HTML]{F8F5FF}\textbf{877} & \cellcolor[HTML]{F8F5FF}\textbf{1424} \\
\cmidrule(lr){2-13}
 & \multirow{2}{*}{LogicBench} & \textbf{Acc. (\%)} & 77.11 & 85.26 & 33.29 & 62.11 & 82.50 & 83.55 & 85.13 & \textbf{83.16} & \textbf{85.00} & \textbf{83.68} \\
 &  & \cellcolor[HTML]{F8F5FF}\textbf{\# Token} & \cellcolor[HTML]{F8F5FF}0 & \cellcolor[HTML]{F8F5FF}1089 & \cellcolor[HTML]{F8F5FF}256 & \cellcolor[HTML]{F8F5FF}256 & \cellcolor[HTML]{F8F5FF}318 & \cellcolor[HTML]{F8F5FF}604 & \cellcolor[HTML]{F8F5FF}816 & \cellcolor[HTML]{F8F5FF}\textbf{256} & \cellcolor[HTML]{F8F5FF}\textbf{512} & \cellcolor[HTML]{F8F5FF}\textbf{438} \\
\cmidrule(lr){2-13}
 & \multirow{2}{*}{TheoremQA} & \textbf{Acc. (\%)} & 52.07 & 68.81 & 14.32 & 28.38 & 56.49 & 62.92 & 66.13 & \textbf{60.24} & \textbf{64.79} & 67.87 \\
 &  & \cellcolor[HTML]{F8F5FF}\textbf{\# Token} & \cellcolor[HTML]{F8F5FF}0 & \cellcolor[HTML]{F8F5FF}935 & \cellcolor[HTML]{F8F5FF}256 & \cellcolor[HTML]{F8F5FF}256 & \cellcolor[HTML]{F8F5FF}281 & \cellcolor[HTML]{F8F5FF}535 & \cellcolor[HTML]{F8F5FF}710 & \cellcolor[HTML]{F8F5FF}\textbf{256} & \cellcolor[HTML]{F8F5FF}\textbf{509} & \cellcolor[HTML]{F8F5FF}727 \\
\cmidrule(lr){2-13}
 & \multirow{2}{*}{ToolQA} & \textbf{Acc. (\%)} & 28.04 & 47.97 & 0.00 & 7.13 & 37.06 & 43.08 & 45.45 & 35.66 & \textbf{43.92} & \textbf{46.92} \\
 &  & \cellcolor[HTML]{F8F5FF}\textbf{\# Token} & \cellcolor[HTML]{F8F5FF}0 & \cellcolor[HTML]{F8F5FF}1010 & \cellcolor[HTML]{F8F5FF}256 & \cellcolor[HTML]{F8F5FF}256 & \cellcolor[HTML]{F8F5FF}294 & \cellcolor[HTML]{F8F5FF}573 & \cellcolor[HTML]{F8F5FF}769 & \cellcolor[HTML]{F8F5FF}256 & \cellcolor[HTML]{F8F5FF}\textbf{512} & \cellcolor[HTML]{F8F5FF}\textbf{754} \\
\midrule
\multirow{10}{*}{\rotatebox[origin=c]{90}{Phi-4}} & \multirow{2}{*}{BigCodeBench} & \textbf{Acc. (\%)} & 43.77 & 52.19 & 0.09 & 0.88 & 45.26 & 47.46 & 49.30 & \textbf{46.58} & \textbf{49.12} & \textbf{50.35} \\
 &  & \cellcolor[HTML]{F8F5FF}\textbf{\# Token} & \cellcolor[HTML]{F8F5FF}0 & \cellcolor[HTML]{F8F5FF}4628 & \cellcolor[HTML]{F8F5FF}256 & \cellcolor[HTML]{F8F5FF}256 & \cellcolor[HTML]{F8F5FF}1284 & \cellcolor[HTML]{F8F5FF}2615 & \cellcolor[HTML]{F8F5FF}3316 & \cellcolor[HTML]{F8F5FF}\textbf{708} & \cellcolor[HTML]{F8F5FF}\textbf{1414} & \cellcolor[HTML]{F8F5FF}\textbf{1889} \\
\cmidrule(lr){2-13}
 & \multirow{2}{*}{CHAMP} & \textbf{Acc. (\%)} & 52.02 & 61.43 & 12.11 & 30.94 & 59.19 & 62.78 & 64.57 & 58.74 & 57.85 & 61.88 \\
 &  & \cellcolor[HTML]{F8F5FF}\textbf{\# Token} & \cellcolor[HTML]{F8F5FF}0 & \cellcolor[HTML]{F8F5FF}1933 & \cellcolor[HTML]{F8F5FF}256 & \cellcolor[HTML]{F8F5FF}256 & \cellcolor[HTML]{F8F5FF}589 & \cellcolor[HTML]{F8F5FF}1114 & \cellcolor[HTML]{F8F5FF}1469 & \cellcolor[HTML]{F8F5FF}438 & \cellcolor[HTML]{F8F5FF}877 & \cellcolor[HTML]{F8F5FF}1664 \\
\cmidrule(lr){2-13}
 & \multirow{2}{*}{LogicBench} & \textbf{Acc. (\%)} & 48.55 & 75.66 & 60.53 & 60.66 & 53.42 & 63.16 & 68.03 & \textbf{63.16} & \textbf{74.74} & \textbf{66.97} \\
 &  & \cellcolor[HTML]{F8F5FF}\textbf{\# Token} & \cellcolor[HTML]{F8F5FF}0 & \cellcolor[HTML]{F8F5FF}1093 & \cellcolor[HTML]{F8F5FF}256 & \cellcolor[HTML]{F8F5FF}256 & \cellcolor[HTML]{F8F5FF}319 & \cellcolor[HTML]{F8F5FF}607 & \cellcolor[HTML]{F8F5FF}820 & \cellcolor[HTML]{F8F5FF}\textbf{256} & \cellcolor[HTML]{F8F5FF}\textbf{512} & \cellcolor[HTML]{F8F5FF}\textbf{335} \\
\cmidrule(lr){2-13}
 & \multirow{2}{*}{TheoremQA} & \textbf{Acc. (\%)} & 45.38 & 51.14 & 15.13 & 18.61 & 47.39 & 45.65 & 48.73 & 43.78 & \textbf{48.33} & \textbf{48.73} \\
 &  & \cellcolor[HTML]{F8F5FF}\textbf{\# Token} & \cellcolor[HTML]{F8F5FF}0 & \cellcolor[HTML]{F8F5FF}921 & \cellcolor[HTML]{F8F5FF}256 & \cellcolor[HTML]{F8F5FF}256 & \cellcolor[HTML]{F8F5FF}272 & \cellcolor[HTML]{F8F5FF}521 & \cellcolor[HTML]{F8F5FF}693 & \cellcolor[HTML]{F8F5FF}256 & \cellcolor[HTML]{F8F5FF}\textbf{509} & \cellcolor[HTML]{F8F5FF}\textbf{682} \\
\cmidrule(lr){2-13}
 & \multirow{2}{*}{ToolQA} & \textbf{Acc. (\%)} & 14.83 & 9.79 & 0.00 & 0.00 & 4.48 & 14.48 & 15.24 & \textbf{23.08} & \textbf{26.78} & \textbf{17.27} \\
 &  & \cellcolor[HTML]{F8F5FF}\textbf{\# Token} & \cellcolor[HTML]{F8F5FF}0 & \cellcolor[HTML]{F8F5FF}996 & \cellcolor[HTML]{F8F5FF}256 & \cellcolor[HTML]{F8F5FF}256 & \cellcolor[HTML]{F8F5FF}287 & \cellcolor[HTML]{F8F5FF}561 & \cellcolor[HTML]{F8F5FF}755 & \cellcolor[HTML]{F8F5FF}\textbf{256} & \cellcolor[HTML]{F8F5FF}\textbf{512} & \cellcolor[HTML]{F8F5FF}\textbf{705} \\
\bottomrule
\end{tabular}%
}
\endgroup
\end{table*}

	We evaluate \textsc{SKIM} on five datasets: BigCodeBench \citep{bigcodebench}, CHAMP \citep{champ}, LogicBench \citep{logicbench}, TheoremQA \citep{theoremqa}, and ToolQA \citep{toolqa}.
	For each task, we use the golden skill annotations provided by SRA-Bench \citep{su2026skillretrievalaugmentationagentic}.
	This setup fixes the relevant skill content for each instance, so the comparison focuses on how the skill is represented in the context.
	These tasks cover code generation, mathematical reasoning, logical reasoning, theorem question answering, and tool use question answering.
	The first four datasets use single-turn QA evaluation, whereas ToolQA uses ReAct multi-turn decision making with tool observations.
	For each instance, the user query remains plain text, while the skill content is either inserted in full or replaced by compressed tokens.
	Table \ref{tab:dataset_statistics} lists the dataset statistics.
	
	\subsection{Models}
	
	We evaluate two target LLMs: Qwen3-8B \citep{qwen3} (with itself as compressor) and the 14B Phi-4 \citep{abdin2024phi4technicalreport} (with the 3.8B Phi-4-mini-instruct as compressor).
	Therefore, this setting lets us examine changes across model families, target model sizes, and compressor sizes.
	For each target model, \textsc{SKIM} runs model specific offline resolution selection for each skill.

	\subsection{Baselines}
	
	We compare against two non compression references.
	Naive does not load any skill, and Full Text loads the golden skills without compression.
	We also include LLMLingua-2 \citep{pan-etal-2024-llmlingua}, a hard token compression baseline which has three settings in the main table.
	The small and medium settings use compression ratios 0.3 and 0.55.
	The large setting uses 0.7 for BigCodeBench and 0.75 for the other tasks, so that its token counts roughly match the higher budget \textsc{SKIM} settings.
	In addition, we report ICAE \citep{icae} and 500xCompressor \citep{li-etal-2025-500xcompressor} as soft compression baselines.
	Since these methods struggle to concatenate multi-span soft tokens for QA tasks, we address multi-skill scenarios by first concatenating the skill texts and then compressing them to a fixed length.
	The selection of these baselines corresponds to those in Table \ref{tab:related_work_comparison} featuring both offline and lightweight compression, as this preserves the inherent nature of skills.
	Further details on baseline implementation are in Appendix \ref{app:experimental_details}.

	\subsection{Implementation Details}
	
	For \textsc{SKIM}, we report three variants.
	\textsc{SKIM} Fix-256 and \textsc{SKIM} Fix-512 use a fixed soft token budget for every skill.
	For the few skills shorter than their compressed form, we directly use the full-text form.
	\textsc{SKIM} Adaptive uses the offline resolution selection to choose among 256 tokens, 512 tokens, and the full text for each skill.
	The offline exam generates 10 diagnostic questions per skill.
	The selection accuracy threshold is 0.9.
	For multi-skill instances, we use the highest selected budget among the referenced skills when a shared compression setting is needed.
	Further hyperparameters and settings, as well as the prompt templates, are provided in Appendix \ref{app:experimental_details} and \ref{app:prompt_templates}, respectively.

	\section{Experimental Results}
	\label{sec:experimental_results}
	
	\subsection{Main Results}

\begin{figure}[!t]
		\centering
		\includegraphics[width=0.9\columnwidth]{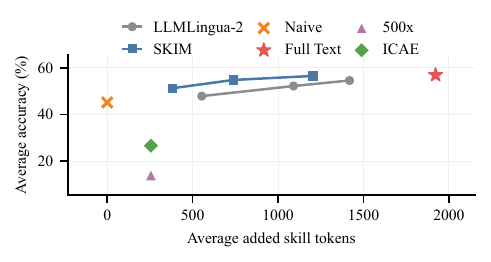}
		\caption{Average token accuracy tradeoff across the two target models and five datasets.
			For LLMLingua-2 and \textsc{SKIM}, settings are first sorted by added skill tokens within each model dataset pair, and the same rank is then averaged across the ten pairs.
			Lines connect settings from the same method family.}
		\label{fig:main_results_tradeoff}
	\end{figure}

	Table \ref{tab:main_results} reports task accuracy and average added tokens across both target models and five datasets.
	Full Text usually outperforms Naive, demonstrating the value of skill annotations, although it also adds hundreds to thousands of tokens per instance.
	Generic soft compression struggles to retain this benefit: ICAE and 500xCompressor often underperform Naive, indicating that factual compression methods transfer poorly to skill settings.
	\textsc{SKIM} achieves a superior accuracy-token tradeoff compared to LLMLingua-2, requiring comparable or smaller token budgets.
	This stems from \textsc{SKIM} aligning continuous tokens with procedural knowledge and employing resolution selection to perceive skill complexity and compression rates, whereas LLMLingua-2 prunes text without explicit procedural execution supervision. Figure \ref{fig:main_results_tradeoff} shows the overall trade-off between accuracy and tokens for all methods, where \textsc{SKIM} performs the best.
	
	Within \textsc{SKIM}, Adaptive improves over fixed budget variants on most datasets, approaching or sometimes exceeding Full Text accuracy while using fewer tokens.
	This follows from the offline exam, which dynamically selects lower resolutions when diagnostic answers are sufficient.
	In a few cases like Phi-4 on LogicBench, a single threshold may heavily favor token reduction by selecting more low-resolution skills, which slightly reduces accuracy relative to Fix-512.
	Figure \ref{fig:qwen_resolution_distribution} in the appendix shows the selected resolution distribution.
	Furthermore, Adaptive uses Full Text answers as silver references. This dependency is usually reasonable, as golden skills generally improve the target model.
	However, if the provided skill text is unhelpful or the model fails to follow it (e.g., Phi-4 on ToolQA, where Full Text underperforms Naive), the resulting weak silver reference causes Adaptive to fall below fixed \textsc{SKIM} variants.
	
	\subsection{Ablation Studies}
	
	Due to the cost of running all variants on every setting, we conduct the ablations mainly with Qwen3-8B on the four datasets excluding ToolQA.
	
	\subsubsection{Training Stages}
	
\begin{figure}[!t]
		\centering
		\includegraphics[width=\columnwidth]{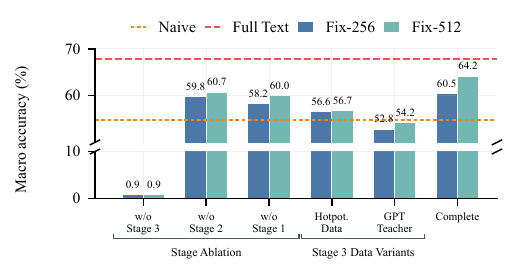}
		\caption{Training ablation for Qwen3-8B over BigCodeBench, CHAMP, LogicBench, and TheoremQA.
			Bars under Stage Ablation remove one training stage.
			Bars under Stage 3 Data Variants replace the Stage 3 skill task data with HotpotQA data or replace self-generated answers with GPT-5.2 teacher answers.
			Complete uses the main \textsc{SKIM} training recipe.}
		\label{fig:training_ablation}
	\end{figure}

	To test whether each training stage contributes to compressed skill use, we compare the full recipe with variants that remove one stage or replace the final skill task data.
	Figure \ref{fig:training_ablation} shows that reconstruction and procedural warm up alone are not sufficient for downstream skill tasks.
	Keeping the skill task stage while removing either earlier stage produces usable performance, but remains below the complete recipe.
	Replacing skill task data with documents and questions from HotpotQA \citep{hotpotqa}, while still using self-generated answers, also lowers performance.
	Replacing the Stage 3 target model answers with GPT-5.2 teacher answers performs poorly.
	This suggests the final stage benefits from target-distribution supervision, rather than from stronger off policy answers alone.

	\subsubsection{Longer Context Stress Test}

\begin{figure}[!t]
		\centering
		\includegraphics[width=0.9\columnwidth]{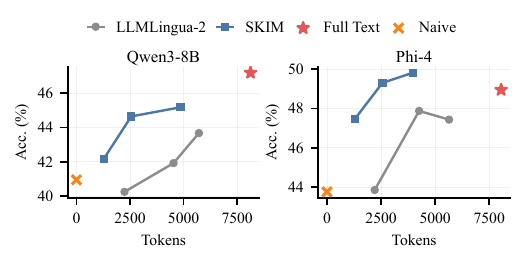}
		\caption{BigCodeBench stress results under retrieved distractor skills.
			LLMLingua-2 and \textsc{SKIM} connect their compression settings.}
		\label{fig:long_context_ablation}
	\end{figure}

	To evaluate retrieval noise and extended contexts, we construct a BigCodeBench stress setting, chosen for its long inputs and outputs.
	To each instance's golden skill set, we appended top-ranked non-gold skills retrieved via \texttt{bge-base-en-v1.5} \citep{bge} until reaching five skills, thereby artificially inflating noise and context length.
	Figure \ref{fig:long_context_ablation} demonstrates that \textsc{SKIM} outperforms LLMLingua-2 at comparable or lower token budgets.
	For Phi-4, higher-resolution \textsc{SKIM} variants surpass Full Text while using fewer tokens.
	This likely occurs because Phi-4's 16k context window is heavily burdened by the uncompressed full text and distractors.
	The results confirm \textsc{SKIM}'s intended benefit: effectively alleviating context pressure while preserving essential skill signals.
	
	
	Additional ablations on offline resolution selection, untrained resolution budgets, and target model adaptation are reported in Appendix \ref{app:further_ablation}.
	
	\section{Conclusion}
	
	In this work, we present \textsc{SKIM}, an adaptive multi-resolution soft token compression framework for procedural skills.
	It trains representations through reconstruction, procedural QA, and skill task alignment, selecting resolutions through offline self-judgment.
	Experiments indicate that \textsc{SKIM} better preserves task accuracy than generic hard and soft baselines under smaller token budgets.
	Thus, \textsc{SKIM} effectively resolves the efficiency-accuracy trade-off in procedural skill deployment.
	
	\section{Limitations}
	
	This work has two limitations.
	First, our experiments use two target LLMs, Qwen3-8B and Phi-4.
	Due to training resource constraints, we do not evaluate \textsc{SKIM} on substantially larger models.
	The results therefore leave open how the accuracy and token tradeoff changes when the target model is stronger.
	Second, \textsc{SKIM} trains model specific projectors and LoRA adapters.
	This design improves alignment between soft tokens and the target LLM, but it also means that a compressed artifact is not directly portable across unrelated model families.
	In settings where a skill repository supports many target models, the repository may need to store several model specific artifacts for the same skill.
	Future work can study whether a unified compressor with target specific projectors can reduce this storage cost while preserving model alignment.

	
	\bibliography{custom}
	
	\appendix
	
	\section{Offline Resolution Selection Algorithm}
	We provide the pseudocode for the offline resolution selection procedure in Algorithm \ref{alg:resolution_selection}.
	\label{app:resolution_selection}

	\begin{algorithm}[t]
		\small
		\caption{Offline resolution selection for one skill}
		\label{alg:resolution_selection}
		\begin{algorithmic}[1]
			\REQUIRE Skill text $s$, resolutions $\mathcal{K}$, question count $N$, threshold $\tau$
			\ENSURE Selected mode for skill $s$
			\STATE Generate questions $\mathcal{Q}=\{q_i\}_{i=1}^{N}$ from $s$
			\FOR{each question $q_i \in \mathcal{Q}$}
			\STATE Generate reference answer $a_i^{\mathrm{full}}$ with the base target model and the full skill text
			\FOR{each resolution $K \in \mathcal{K}$}
			\STATE Generate candidate answer $a_i^K$ with $E_K(s)$
			\STATE Judge whether $a_i^K$ matches $a_i^{\mathrm{full}}$
			\ENDFOR
			\ENDFOR
			\FOR{each resolution $K \in \mathcal{K}$}
			\STATE Compute $\alpha_K(s)$ as the fraction of accepted candidate answers
			\ENDFOR
			\IF{there exists $K$ such that $\alpha_K(s) \geq \tau$}
			\STATE \textbf{return} the smallest such $K$
			\ELSE
			\STATE \textbf{return} full skill text
			\ENDIF
		\end{algorithmic}
	\end{algorithm}

	\section{Stage 3 Data Construction Algorithm}
	\label{app:stage3_data_construction}
	
	Algorithm \ref{alg:stage3_data_construction} summarizes the Stage 3 data construction procedure.
	GPT-5.2 is used only for source skill evaluation, question generation, and skill splitting.
	For each target model, \textsc{SKIM} runs answer generation with that target model.

	\section{Dataset and Hyperparameter Details}
	\label{app:experimental_details}
	
	Table \ref{tab:dataset_statistics} summarizes the five datasets used in the main comparison.
	The skill columns are computed from the SRA-Bench golden skill annotations, while the question counts come from the task instances.
	BigCodeBench and CHAMP contain multi-skill cases, whereas LogicBench, TheoremQA, and ToolQA use a single annotated skill per instance.
	ToolQA is the only dataset in this group that requires tool style interaction.
	We do not include MedCalcBench \citep{medcalcbench} in the main comparison, since some of its skills are short and do not provide enough room to evaluate compression behavior.
	
	\begin{algorithm}[t]
		\small
		\caption{Stage 3 skill task data construction}
		\label{alg:stage3_data_construction}
		\begin{algorithmic}[1]
			\REQUIRE Source skills $\mathcal{S}$, target model $M$, evaluator $G$
			\ENSURE Stage 3 training set $\mathcal{D}$ for $M$
			\STATE Notation: $v$ is validity, $\mathcal{Q}$ questions, $\mathcal{T}$ tool specs, $b_{\mathrm{split}}$ split flag, $b_{\mathrm{react}}$ ReAct flag
			\STATE Notation: $\mathcal{R}$ is skill context, $\tau$ trace, $q$ question, $a$ answer, $h$ thought, $u$ action, $o$ observation
			\STATE $\mathcal{D} \leftarrow \emptyset$
			\FOR{each source skill $s \in \mathcal{S}$}
			\STATE $(v,\mathcal{Q},\mathcal{T},b_{\mathrm{split}},b_{\mathrm{react}}) \leftarrow \mathrm{EvalSkill}_{G}(s)$
			\IF{$v=\mathrm{false}$}
			\STATE continue
			\ENDIF
			\IF{$b_{\mathrm{split}}=\mathrm{true}$}
			\STATE $\mathcal{R} \leftarrow \mathrm{SplitSkill}_{G}(s)$
			\ELSE
			\STATE $\mathcal{R} \leftarrow \{s\}$
			\ENDIF
			\FOR{each question $q \in \mathcal{Q}$}
			\IF{$b_{\mathrm{react}}=\mathrm{false}$}
			\STATE $a \leftarrow \mathrm{DirectAnswer}_{M}(\mathcal{R},q)$
			\STATE Add $(\mathcal{R},q,a)$ to $\mathcal{D}$
			\ELSE
			\STATE $\tau \leftarrow \emptyset$, $\mathrm{finished} \leftarrow \mathrm{false}$
			\WHILE{$\mathrm{finished}=\mathrm{false}$}
			\STATE $(h,u) \leftarrow \mathrm{ReactStep}_{M}(\mathcal{R},q,\mathcal{T},\tau)$
			\IF{$u$ is \textsc{Finish}}
			\STATE Append $(h,u)$ to $\tau$ and set $\mathrm{finished} \leftarrow \mathrm{true}$
			\ELSE
			\STATE $o \leftarrow \mathrm{SimulateTool}_{M}(q,\mathcal{T},u,\tau)$
			\STATE Append $(h,u,o)$ to $\tau$
			\ENDIF
			\ENDWHILE
			\STATE Add $(\mathcal{R},q,\tau)$ to $\mathcal{D}$
			\ENDIF
			\ENDFOR
			\ENDFOR
			\STATE \textbf{return} $\mathcal{D}$
		\end{algorithmic}
	\end{algorithm}
	
	\begin{table*}[!t]
\centering
\small
\caption{Task statistics for the five evaluation datasets.
Question counts follow the task data, and gold skill counts follow the corresponding skill annotations.
Avg. skills reports the mean number of annotated gold skills per test instance.
Mixed means that both single skill and multi skill instances appear.}
\label{tab:dataset_statistics}
\resizebox{0.95\textwidth}{!}{%
\begin{tabular}{llrrrcc}
\toprule
\textbf{Dataset} & \textbf{Task Type} & \textbf{Questions} & \textbf{Gold Skills} & \textbf{Avg. Skills} & \textbf{Multi Skill} & \textbf{Tool Use} \\
\midrule
BigCodeBench & Code generation & 1,140 & 139 & 2.76 & Yes & No \\
CHAMP & Math reasoning & 223 & 89 & 1.71 & Mixed & No \\
LogicBench & Logical reasoning & 760 & 19 & 1.00 & No & No \\
TheoremQA & Theorem question answering & 747 & 320 & 1.00 & No & No \\
ToolQA & Tool use question answering & 1,430 & 14 & 1.00 & No & Yes \\
\bottomrule
\end{tabular}%
}
\end{table*}

	Next, we show the training data details of \textsc{SKIM}. For Stage 1, we use an open-source skill dataset, where skills are collected from GitHub repositories.\footnote[2]{https://huggingface.co/datasets/LittleDinoC/agent-skills}
	We filter out skills shorter than 500 characters, since these examples are often low quality or underspecified.
	We also verify that the minimum character edit distance between any retained training skill and any golden skill in the test dataset is greater than 2200, which indicates that the evaluation golden skills do not appear in the training pool.
	Table \ref{tab:training_data_scale} reports the approximate size of the data used by each \textsc{SKIM} training stage.
	Stage 1 uses the filtered collected skill documents for reconstruction, Stage 2 uses WikiHow procedural QA, and Stage 3 uses generated skill task QA from evaluated source skills. During the training stage, we employ DeepSpeed \citep{deepspeed} to improve training efficiency and reduce memory consumption.

\begin{table*}[!t]
\centering
\small
\caption{Approximate training data scale for \textsc{SKIM}.}
\label{tab:training_data_scale}
\resizebox{0.95\textwidth}{!}{%
\begin{tabular}{lll}
\toprule
\textbf{Stage} & \textbf{Data source} & \textbf{Approximate scale} \\
\midrule
Stage 1 & Collected skill documents & 60.9k training skills \\
Stage 2 & WikiHow procedural QA & 214.3k processed QA examples from WikiHow, using the cleaned version\footnotemark[4] \\
Stage 3 & Generated skill task QA & Evaluated source skills with 5 candidate questions per skill, totally \textasciitilde 60k training examples \\
\bottomrule
\end{tabular}%
}
\end{table*}

	Table \ref{tab:training_hyperparameters} lists the main \textsc{SKIM} implementation settings.
	It includes the model and compressor pairs, candidate resolution budgets, projector configuration, LoRA settings, offline exam parameters, and inference decoding parameters.
	
	\begin{table*}[!t]
\centering
\small
\caption{Implementation details for \textsc{SKIM}.}
\label{tab:training_hyperparameters}
\resizebox{0.95\textwidth}{!}{%
\begin{tabular}{lll}
\toprule
\textbf{Group} & \textbf{Setting} & \textbf{Value} \\
\midrule
Models & Qwen3-8B target and compressor & Qwen3-8B target, Qwen3-8B compressor \\
Models & Phi-4 target and compressor & Phi-4 14B target, Phi-4-mini-instruct 3.8B compressor \\
Resolution & Candidate budgets & 256 and 512 soft tokens, with full text fallback \\
Architecture & Projector & 3 layers, hidden size 8192 \\
Architecture & Low resolution construction & Prefix truncation from the 512 token artifact \\
Training & Shared settings & Weight decay 0.01, warmup ratio 0.02, bfloat16, gradient checkpointing \\
Training & Stage 1 & Learning rate $1 \times 10^{-4}$, global batch size 192, max length 2048 \\
Training & Later stages & Learning rate $7 \times 10^{-5}$, global batch size 192, max length 2048 \\
Training & Stage 1 schedule & At most 20 epochs, with early stopping when training loss is below $2 \times 10^{-3}$ \\
Training & Stage 2 schedule & 4 epochs as warm up \\
Training & Stage 3 schedule & Early stopping when training loss is below $0.25$ \\
LoRA & Target LLM adaptation & Rank 16, alpha 32, dropout 0.05 \\
Offline exam & Question generation & 10 questions per skill, temperature 0.2, max output 2048 tokens \\
Offline exam & Judgment and selection & Full text answer as reference, temperature 0.0, threshold 0.9 \\
Inference & Decoding & No sampling, temperature 0.1, top $p$ 0.95, max new tokens 1024 \\
\bottomrule
\end{tabular}%
}
\end{table*}

	Table \ref{tab:baseline_hyperparameters} summarizes the implementation details for ICAE and 500xCompressor.
	For their training data, we align with ATACompressor \citep{atacompressor} and extract contexts with QA supervision from HotpotQA \citep{hotpotqa} and MS-MARCO \citep{msmarco}.
	We sample about 60,000 training examples, which is comparable to the source pool used for our Stage 3 skill task alignment data. We employ the official 500xCompressor code with the default experimental setup.\footnote[3]{https://github.com/ZongqianLi/500xCompressor}
	
	\begin{table*}[!t]
\centering
\small
\caption{Key hyperparameters for the ICAE and 500xCompressor soft compression baselines.}
\label{tab:baseline_hyperparameters}
\resizebox{0.8\textwidth}{!}{%
\begin{tabular}{lll}
\toprule
\textbf{Group} & \textbf{Setting} & \textbf{Value} \\
\midrule
Data & Source datasets & HotpotQA and MS-MARCO QA records \\
Data & Sampling and split & 60k sampled examples \\
Compression & Memory budget & 256 memory tokens \\
Pretraining & Schedule & 3 epochs, learning rate $1 \times 10^{-4}$, warmup 300 steps \\
QA fine tuning & Schedule & 10 epochs, learning rate $5 \times 10^{-5}$, warmup 300 steps \\
LoRA & Adapter configuration & Rank 64, alpha 32, dropout 0.05 \\
\bottomrule
\end{tabular}%
}
\end{table*}

	\footnotetext[4]{https://huggingface.co/datasets/gursi26/wikihow-cleaned}

	\section{Prompt Templates}
	\label{app:prompt_templates}
	
	This appendix lists the prompt templates used by the data construction and offline exam stages.
	The skill evaluation prompt is used before Stage 3 to filter collected skills, generate candidate user questions, identify tools, and decide whether ReAct style answers are appropriate.
	The skill decomposition prompt is used in the same data preparation stage to split a complex skill into self contained subskills, which gives controlled multi-skill training examples.
	The skill QA answer guidance prompt is appended when the target model generates supervised answers for Stage 3, so that answer generation follows the provided skill instead of relying only on generic model behavior.
	The tool observation simulator prompt is used during ReAct data generation to produce Observation text from a tool schema, an action, and the partial trajectory.
	The offline question generation and answer judgment prompts are used by the resolution selection exam described in Algorithm \ref{alg:resolution_selection}.
	
	\begin{tcolorbox}[breakable,colback=lightgray!20,colframe=darkgray!80,title=Skill Evaluation Prompt]
\small
\textbf{System Prompt.}
You are an expert evaluator for AI agent skills. Return only valid JSON. Do not output markdown. Follow the schema exactly.
\\ \hspace*{\fill} \\ \relax
\textbf{User Prompt Template.}
You are given one skill object in JSON. Evaluate this skill from the perspective of whether it provides useful additional constraints, operational guidance, or capabilities to a base LLM.
\\ \hspace*{\fill} \\ \relax
Tasks:
0) Give quality\_score from 0 to 5 (higher is better).
1) Generate exactly \{k\} concrete user questions that this skill can solve. These questions should be the users' actual real-life problems, rather than merely asking about the content of the Skill.
2) Estimate whether this skill can be split into independent sub-skills that each solve one problem without depending on other skills. Return split\_skill\_count (0 if not splittable).
3) Identify tools involved in this skill. If no tools are needed, return an empty list. If tools are involved, return a list where each tool includes: name, description, input params (name + description), output params (name + description). Environment which returns information can be treated as tools too. The name of the tool is a function name that starts with a capital letter.
4) Judge whether the problem-solving flow is better suited to ReAct (Thought-Action-Observation). Return a boolean and a short reason.
\\ \hspace*{\fill} \\ \relax
Output requirements:
- Return valid JSON only.
- Keep field names exactly as provided in the schema.
- No additional top-level fields beyond the schema.
\\ \hspace*{\fill} \\ \relax
Schema:
\{output\_schema\}
\\ \hspace*{\fill} \\ \relax
Skill JSON:
\{skill\_json\}
\end{tcolorbox}

\begin{tcolorbox}[breakable,colback=lightgray!20,colframe=darkgray!80,title=Skill Decomposition Prompt]
\small
\textbf{System Prompt.}
You are an expert skill decomposition assistant for AI agent skills. Return only valid JSON. Do not output markdown code fences.
\\ \hspace*{\fill} \\ \relax
\textbf{User Prompt Template.}
You are given one original skill and its evaluation result. Decompose it into exactly \{split\_count\} independent sub-skills. Each sub-skill must be self-contained and able to solve one clear problem without relying on other skills.
\\ \hspace*{\fill} \\ \relax
Hard requirements:
1) Output exactly \{split\_count\} sub-skills.
2) Every sub-skill must contain all required fields: name, description, content.
3) content must be markdown text and should be practical, executable skill content (not placeholder text).
4) Preserve essential operational details from the original skill that are needed to solve the target problem.
5) Avoid overlap between sub-skills as much as possible.
\\ \hspace*{\fill} \\ \relax
Output requirements:
- Return valid JSON only.
- Keep top-level keys exactly as in the schema.
- No additional top-level keys.
\\ \hspace*{\fill} \\ \relax
Schema:
\{output\_schema\}
\\ \hspace*{\fill} \\ \relax
Input skill JSON:
\{skill\_json\}
\end{tcolorbox}

\begin{tcolorbox}[breakable,colback=lightgray!20,colframe=darkgray!80,title=Skill QA Answer Guidance Prompt]
\small
\textbf{Generation Add Extra Prompt.}
\\ \hspace*{\fill} \\ \relax
\\ \hspace*{\fill} \\ \relax
Responses must follow the suggestions or guidelines provided in the Skill text. Seamlessly integrate the Skill text guidelines into your problem-solving process. Justify your planning steps as they occur by explicitly linking them to the relevant key content from the Skill text.
\end{tcolorbox}

\begin{tcolorbox}[breakable,colback=lightgray!20,colframe=darkgray!80,title=Tool Observation Simulator Prompt]
\small
\textbf{Tool Simulator System Prompt.}
You simulate tool execution for ReAct training data. Always return a concrete best-effort tool output as plain text. Never refuse, never say you do not know, never say mock data, and never return placeholders.
\\ \hspace*{\fill} \\ \relax
\textbf{Tool Simulator User Template.}
You are simulating a tool call in a QA ReAct trajectory.
\\ \hspace*{\fill} \\ \relax
Question:
\{query\}
\\ \hspace*{\fill} \\ \relax
Available tools (JSON):
\{tools\_json\}
\\ \hspace*{\fill} \\ \relax
Current action:
\{action\}
Action name: \{action\_type\}
Action arguments: \{action\_args\}
\\ \hspace*{\fill} \\ \relax
Conversation scratchpad (truncated):
\{scratchpad\}
\\ \hspace*{\fill} \\ \relax
Return exactly the observation text for this action. It must be specific and plausible.
\end{tcolorbox}

\begin{tcolorbox}[breakable,colback=lightgray!20,colframe=darkgray!80,title=Offline Question Generation Prompt]
\small
\textbf{System Prompt.}
You are an expert evaluator that designs benchmark QA questions for skill capability testing. Return valid JSON only.
\\ \hspace*{\fill} \\ \relax
\textbf{User Prompt Template.}
You are given one skill object. Generate exactly \{n\_questions\} user questions that can be solved with this skill.
\\ \hspace*{\fill} \\ \relax
Requirements:
1) Questions must be concrete, answerable, and diverse.
2) Questions should focus on practical task completion, not asking to summarize the skill itself.
3) Questions must test mastery of the skill WITHOUT leaking the skill content directly. Do not ask questions that can be answered by simply copying from the skill text.
4) If the skill involves code/programming or tool usage, questions should focus on code generation or tool calling tasks that require applying the skill's knowledge.
5) Keep each question concise but specific.
6) Return valid JSON only with exactly the schema below.
\\ \hspace*{\fill} \\ \relax
Schema:
\{output\_schema\}
\\ \hspace*{\fill} \\ \relax
Skill JSON:
\{skill\_json\}
\end{tcolorbox}

\begin{tcolorbox}[breakable,colback=lightgray!20,colframe=darkgray!80,title=Offline Answer Judgment Prompt]
\small
\textbf{System Prompt.}
You are a strict answer equivalence judge. Return valid JSON only.
\\ \hspace*{\fill} \\ \relax
\textbf{User Prompt Template.}
Judge whether the candidate answer is correct compared to the reference answer for the same question.
\\ \hspace*{\fill} \\ \relax
Dataset style: \{dataset\}
Question:
\{question\}
\\ \hspace*{\fill} \\ \relax
Reference answer (ground truth from full\_text mode):
\{reference\_answer\}
\\ \hspace*{\fill} \\ \relax
Candidate answer:
\{candidate\_answer\}
\\ \hspace*{\fill} \\ \relax
Rules:
1) Judge semantic correctness, not string identity.
2) If candidate contradicts key facts in reference, mark incorrect.
3) If reference is empty/invalid, this item should be skipped by caller (not judged).
4) Output valid JSON only with schema below.
\\ \hspace*{\fill} \\ \relax
Schema:
\{output\_schema\}
\end{tcolorbox}

	\begin{figure}[!t]
		\centering
		\includegraphics[width=\columnwidth]{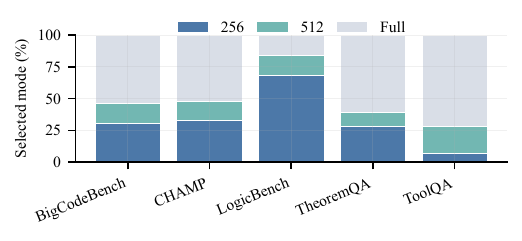}
		\caption{Distribution of final \textsc{SKIM} resolution choices for Qwen3-8B across the datasets in Table \ref{tab:main_results}.
			Bars show the proportion assigned to 256 tokens, 512 tokens, or Full Text by the offline exam procedure.}
		\label{fig:qwen_resolution_distribution}
	\end{figure}
	
	\section{Further Ablation Studies}
	\label{app:further_ablation}
	
	\subsection{Offline Resolution Selection}
	\label{app:offline_resolution_selection_ablation}
	
\begin{figure*}[!t]
		\centering
		\includegraphics[width=0.95\textwidth]{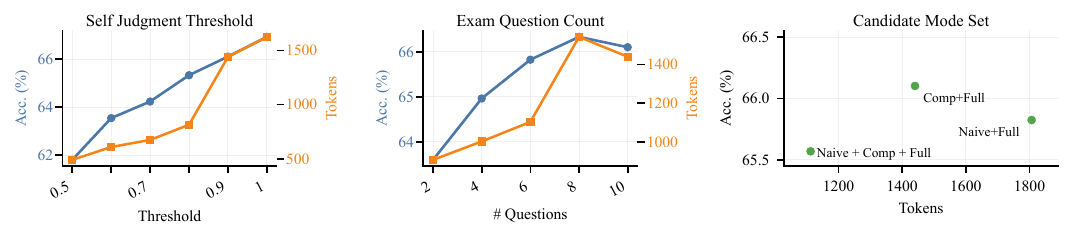}
		\caption{Offline resolution selection ablations for Qwen3-8B on BigCodeBench, CHAMP, LogicBench, and TheoremQA.
			The first panel varies the self judgment accuracy threshold, the second varies the number of generated exam questions per skill, and the third varies the candidate modes available during the exam.
			The first two panels report macro accuracy and macro average added tokens, while the third plots candidate settings by token count and accuracy.
			In the third panel, Comp+Full uses compressed and full skill candidates, Naive+Full uses the no skill answer and full skill text, and Naive + Comp + Full enables all three candidate types.}
		\label{fig:resolution_selection_ablation}
	\end{figure*}

	To test whether the offline exam controls the accuracy and token tradeoff, we vary the judgment threshold, the number of generated questions, and the candidate modes.
	Figure \ref{fig:resolution_selection_ablation} shows the expected pattern in which stricter thresholds load more skill information and improve accuracy.
	The main threshold of 0.9 is close to the strictest setting in accuracy while using fewer tokens.
	Generating more exam questions also improves selection quality, at the cost of selecting larger budgets more often.
	Finally, allowing the exam to choose among compressed candidates and Full Text gives the best accuracy among the tested candidate sets.
	Adding Naive can reduce tokens, but it also lowers accuracy, which suggests that the exam is more reliable when it decides how much skill information to load instead of deciding whether to load any skill information.
	Moreover, applying the exam without compressed candidates uses more tokens than our main setting and still gives lower accuracy.

	\begin{figure}[!t]
		\centering
		\includegraphics[width=\columnwidth]{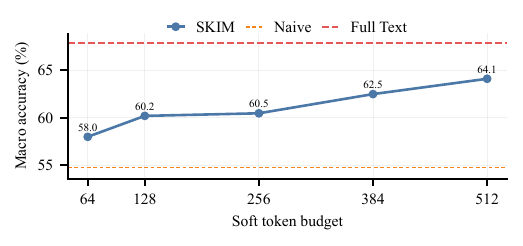}
		\caption{Untrained resolution budget ablation for Qwen3-8B on BigCodeBench, CHAMP, LogicBench, and TheoremQA.
			Only 256 and 512 tokens are used as training resolutions, while 64, 128, and 384 tokens are evaluated by prefix truncation at inference time.
			Dashed lines mark the Naive and Full Text references.}
		\label{fig:untrained_resolution_ablation}
	\end{figure}

	\begin{figure}[!t]
		\centering
		\includegraphics[width=\columnwidth]{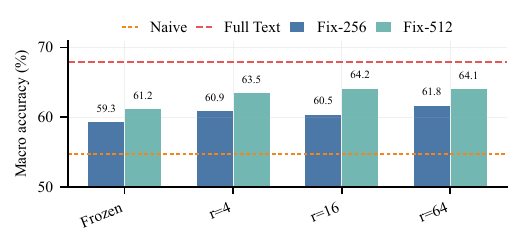}
		\caption{Third stage LoRA ablation for Qwen3-8B on BigCodeBench, CHAMP, LogicBench, and TheoremQA.
			This ablation is run in the skill task alignment stage.
			Frozen trains the compressor side while keeping the target LLM weights fixed, and the rank variants use LoRA with alpha set to twice the rank.}
		\label{fig:lora_ablation}
	\end{figure}
	
	\subsection{Untrained Resolution Budgets}
	\label{app:untrained_resolution_budgets}
	
	To test whether the multi-resolution representation transfers beyond the budgets used during training, we evaluate Qwen3-8B with soft token budgets $K \in \{64,128,256,384,512\}$.
	Only 256 and 512 tokens are used as training resolutions, while the other budgets are obtained by prefix truncation at inference time.
	Figure \ref{fig:untrained_resolution_ablation} shows that the model remains above the Naive reference even at 64 tokens, and accuracy generally increases as the budget grows.
	The 384 token setting is not a training resolution, but it improves over 256 tokens and approaches the 512 token result.
	This pattern suggests that \textsc{SKIM} learns a representation that can adapt to intermediate and lower budgets, although Full Text still provides the upper reference.

	\subsection{Target Model Adaptation}
	
	To test whether the target model must adapt to continuous skill tokens, we compare frozen target LLM training with several LoRA ranks in the third stage.
	Figure \ref{fig:lora_ablation} shows that freezing the target LLM is weaker than using LoRA, especially at the larger resolution.
	Small rank LoRA already recovers most of the gain, and larger ranks are close to each other.
	Thus, target model adaptation is important, but excessively large LoRA ranks are unnecessary.
	
	\section{Qualitative Case Study}
	\label{app:qualitative_case}
	
	Table \ref{tab:toolqa_case} shows a ToolQA coffee lookup case from \texttt{toolqa\_008}.
	The factual schema question is answered by both compressed methods, while the procedural lookup requires the model to preserve the ordered database operations specified by the skill.
	
	\section{Licensing}
	Qwen3-8B is released under the
	Apache License 2.0. Phi-4, Phi-4-mini-Instruct, and bge-base-en-v1.5 are
	released under the MIT license.
	For the datasets, LogicBench, TheoremQA, CHAMP, and SRA-Bench are released under the MIT license.
	BigCodeBench and ToolQA are released under the Apache License 2.0.
	The objective of this paper is academic exploration, which is consistent with the permitted use under this license.

	\begin{table*}[!t]
		\centering
		\small
		\renewcommand{\arraystretch}{1.18}
		\resizebox{\textwidth}{!}{
			\begin{tabular}{@{}p{0.96\textwidth}@{}}
				\toprule
				\textbf{Skill Content.}
				\texttt{\# Coffee Price Data, Single Day Lookup Guide}
				\newline
				\texttt{\#\# Data Source}
				\newline
				The coffee database (\texttt{LoadDB[coffee]}) contains daily coffee futures price data.
				Columns: \texttt{Date}, \texttt{Open}, \texttt{High}, \texttt{Low}, \texttt{Close}, \texttt{Volume}, \texttt{Currency}.
				\newline
				Dates are stored in YYYY-MM-DD format.
				\newline
				All values are stored as strings internally.
				\newline
				\texttt{\#\# Standard Lookup Workflow}
				\newline
				For any question about a specific trading day, use this three step workflow:
				\newline
				\texttt{1. LoadDB[coffee]}
				\newline
				\{... The remaining skill specifies the exact single day lookup workflow, column semantics, derived computations, and notes for missing trading days.\} \\
				\midrule
				\textbf{Factual question.} What columns does this database contain?
				\newline
				\textbf{Answer from ICAE.} \textcolor{red}{Date, Open, High, Low, Close, Volume, Currency}
				\newline
				\vspace{0.35em}
				\newline
				\textbf{Answer from \textsc{SKIM}.} The coffee database contains the following columns:
				\textcolor{red}{Date}, \textcolor{red}{Open}, \textcolor{red}{High}, \textcolor{red}{Low}, \textcolor{red}{Close}, \textcolor{red}{Volume}, and \textcolor{red}{Currency}. \\
				\midrule
				\textbf{Procedural question.} What was the opening price of coffee on 2001-12-06?
				\newline
				\textbf{Answer from ICAE.}
				Thought 1: 123.00
				\newline
				Action 1: Finish[123.00]
				\newline
				Observation 1: Answer: 123.00
				\newline
				\vspace{0.35em}
				\newline
				\textbf{Answer from \textsc{SKIM}.}
				Thought 1: This is a question related to coffee prices. We need to load the coffee database.
				\newline
				Action 1: \textcolor{red}{LoadDB[coffee]}
				\newline
				Observation 1: We have successfully loaded the coffee database, including the following columns: Date, Open, High, Low, Close, Volume, Currency.
				\newline
				Thought 2: We need to filter the information related to the query.
				\newline
				Action 2: \textcolor{red}{FilterDB[Date=2001-12-06]}
				\newline
				Observation 2: We have successfully filtered the data (1 rows).
				\newline
				Thought 3: We then need to know the opening price.
				\newline
				Action 3: \textcolor{red}{GetValue[Open]}
				\newline
				Observation 3: \textcolor{red}{42.75}
				\newline
				Thought 4: The opening price of coffee on 2001-12-06 is \textcolor{red}{42.75}.
				\newline
				Action 4: \textcolor{red}{Finish[42.75]}
				\newline
				Observation 4: Answer: \textcolor{red}{42.75} \\
				\bottomrule
		\end{tabular}}
		\caption{Qualitative comparison on a ToolQA coffee database skill.
			Red text marks factual fields or procedural steps that match the expected lookup behavior.
			In this case, both methods recover the factual schema, but only \textsc{SKIM} preserves the operation sequence needed to answer the value lookup.}
		\label{tab:toolqa_case}
	\end{table*}
	
\end{document}